# BioLORD-2023: Semantic Textual Representations Fusing LLM and Clinical Knowledge Graph Insights

François Remy, Kris Demuynck, Thomas Demeester [Ghent University - imec]

## Abstract

### Objective

In this study, we investigate the potential of Large Language Models to complement biomedical knowledge graphs in the training of semantic models for the biomedical and clinical domains.

### Materials and Methods

Drawing on the wealth of the UMLS knowledge graph and harnessing cutting-edge Large Language Models, we propose a new state-of-the-art approach for obtaining high-fidelity representations of biomedical concepts and sentences, consisting of three steps: an improved contrastive learning phase, a novel self-distillation phase, and a weight averaging phase.

### Results

Through rigorous evaluations via the extensive BioLORD testing suite and diverse downstream tasks, we demonstrate consistent and substantial performance improvements over the previous state of the art (e.g. +2pts on MedSTS, +2.5pts on MedNLI-S, +6.1pts on EHR-Rel-B). Besides our new state-of-the-art biomedical model for English, we also distill and release a multilingual model compatible with 50+ languages and finetuned on 7 European languages.

### Discussion

Many clinical pipelines can benefit from our latest models. Our new multilingual model enables a range of languages to benefit from our advancements in biomedical semantic representation learning, opening a new avenue for bioinformatics researchers around the world. As a result, we hope to see BioLORD-2023 becoming a precious tool for future biomedical applications.

### Conclusion

In this paper, we introduced BioLORD-2023, a state-of-the-art model for semantic textual similarity and biomedical concept representation designed for the clinical domain.

# INTRODUCTION

Clinical and Biomedical Natural Language Processing (Clinical NLP) rose to prominence in the last few years [1], debuting a long series of surveys [2-8] highlighting the potential and inherent challenges of harnessing the synergies between, on one hand, the reliable insights originating from biomedical knowledge graphs and, on the other hand, the impressive generalization capabilities of cutting-edge deep learning techniques, specifically large language models (LLMs).

Such an insight fusion could have an immense impact across a wide range of applications, encompassing clinical case summarization, clinical decision support, patient diagnosis and triage, pharmacovigilance, disease subtyping, drug discovery, as well as help researchers build more explainable AI systems.

In this work, we continue to pave the trail previously blazed by BioSyn [9], SapBERT [10] and BioLORD [11], by integrating knowledge graph information during and after the pre-training of semantic bidirectional language models [12]. These bidirectional models form a cornerstone of the modern Retrieve-And-Generate pipeline (RAG) [13,14], which is critical in most applications of LLMs in a real-world setting (both to enable LLMs to access accurate and up-to-date data before writing their answers, but also because it makes tracing and combating erroneous answers more tractable).

This article introduces several novel contributions to this existing body of work, aiming at <1> broadening the biomedical expertise of semantic models, <2> reducing the trade-off between the biomedical knowledge and the general language understanding of finetuned models, and <3> enabling more languages to benefit from the obtained improvements.

To this end, we present in this article a new model, BioLORD-2023, which builds upon the achievements of the original BioLORD model [11] but has novel characteristics, such as an improved training strategy and an updated training corpus. The original BioLORD will henceforth be referred to as BioLORD-2022, and compared against the new BioLORD-2023, to avoid confusion.

The first contribution of BioLORD-2023 concerns the usage of Large Language Models [15] for the verbalization of the knowledge graph information contained in biomedical ontologies [16,17], and the useful addition of their latent biomedical knowledge to these knowledge verbalization efforts.

Indeed, only 5% of clinical concepts possess human-written definitions in large biomedical meta-thesauri [11,16]. Fortunately, modern LLMs prompted with knowledge graph information have been shown to generate largely reliable, insightful, and fluent definitions for a vast majority of biomedical concepts [18]. Yet, the practical benefits of such LLM-generated definitions have not been studied extensively so far, something we aim to address.

In this paper, we demonstrate that the existence of these artificial definitions for a large majority of biomedical concepts can indeed substantially enhance the quality of representations obtained using the "Learning of Ontological Representations through Definitions and textual representations" strategy (LORD) [11], by including such definitions in the training data of our new model and assessing its impact on downstream tasks.

Our second contribution, a self-distillation approach, takes advantage of the existence of this broad set of definitions to accelerate the convergence process of the LORD training strategy, thereby achieving superior biomedical knowledge acquisition at a reduced loss of general language understanding capabilities.

Combining these two strategies, we train a new biomedical semantic model, BioLORD-2023. We evaluate our newly-trained BioLORD-2023 model on a broad spectrum of downstream tasks, including Biomedical Concept

Representation (BCR), Semantic Textual Similarity (STS), and Named Entity Linking (NEL), with considerable gains across the entire range of tasks.

Our third and last contribution is the release of our first multi-lingual clinical language model, enabling the retrieval and the concept normalization of content in up to 50 languages, thanks to the cross-lingual distillation strategy described by Reimers et al. [19] and a multi-lingual alignment dictionary built from SnomedCT [17] using LaBSE [20].

We evaluate this new multilingual model based on the test suite developed for multilingual-SapBERT [21], a similar model which is widely considered as the current state-of-the-art in the domain. We also evaluate the quality of the distillation process using our evaluation metrics for the English language.

To summarize, our three main contributions are:

1. The expansion of our training corpus by supplementing its existing knowledge with new LLMs-generated definitions for 400,000 concepts, fusing Knowledge Graph and LLM insights inside the LORD pre-training.
2. The introduction of a novel self-distillation technique to speed up biomedical knowledge acquisition while preserving the language understanding capabilities of the BioLORD-type models.
3. The delivery of a state-of-the-art multilingual model for the biomedical domain, using a proven cross-lingual distillation technique.

# RELATED WORKS

Before delving into the details of our methodology, we provide a short description of the context in which this work finds its place, the technologies used in this paper, and the previous efforts which made this work possible, starting with biomedical knowledge graphs.

## Biomedical Knowledge Graphs

Biomedical knowledge graphs (BKGs) are graph-based representations of biomedical data and knowledge, where nodes represent entities (e.g., genes, diseases, drugs) and edges represent relationships (e.g., interactions, associations, causations) between these entities. BKGs can be classified by their scope, curation strategy, and structure, and they can contain various types of information [22]. Several applications and tasks in biomedicine and healthcare have been shown to benefit from BKGs, both because BKGs can be used as reliable sources of information, and because they provide traceable explanations for answers which can be derived from them.

UMLS [16] and SNOMED CT [17] are two examples of BKGs that are used to standardize health and clinical information. They differ from each other both in scope and structure. UMLS, as a conglomerate of biomedical vocabularies, aims for a large coverage and encompasses over 3.7 million concepts from 200+ source vocabularies, including SNOMED CT; however, it does not provide consistent views for all its concepts. Conversely, SNOMED CT aims to provide a reliable gold standard for electronic health record standardization, thanks to its meticulous formal logic-based structure and approximately 358,000 clinical concepts.

In this work, we maximally leverage the strengths of both UMLS and SnomedCT by using them in the learning phases for which they are best suited. For instance, in continuity with previous iterations of BioLORD, we employ the UMLS concepts and relationships as part of the contrastive pre-training, where the increased scope and diversity of described relationships is an advantage. However, we also employ definitions from the Automatic Glossary of Clinical Terminology (AGCT) [18], which makes use of SnomedCT's more standardized and consistently-annotated graph to increase the homogeneity and reliability of the produced definitions.

## Large Language Models in Healthcare

Large Language Models (LLMs) encompass various types of machine learning models which have been trained on vast amounts of text to either achieve some understanding of existing content or generate original content based on instructions. They have recently demonstrated remarkable capabilities in natural language processing tasks and beyond.

Large Language Models (LLMs) have been used in various clinical applications [23-25]. One such application is medical transcription and clinical coding using the international classification of diseases (ICD), where LLMs have been used to improve the accuracy and efficiency of converting spoken medical observations into written or structured electronic health records (EHRs) [26].

LLMs also show promises in various clinical data analysis tasks, for instance analyzing patient data such as medical records [23], or interpreting imaging studies and laboratory results [27,28], thereby providing valuable insights that aid doctors and other healthcare professionals in diagnosing patients. Finally, LLMs can also be used to identify clinical trial opportunities for patients by analyzing patient data such as medical records [29].

Examples of large biomedical language models include Med-PALM [30], Galactica [31], ClinicalGPT [32], BioMedLM [33], BioGPT [34] and others. Commercial language models such as ChatGPT [15] and GPT-4 [35] have also shown great capabilities in Biomedical AI [25], despite lacking dedicated finetuning procedures.

## Retrieval-Augmented Generation

While Large Language Models have demonstrated remarkable capabilities in natural language processing tasks, they are prone to hallucinations, which can result in incorrect diagnoses and treatments, leading to adverse effects on patients. Retrieval-Augmented Generation (RAG) is a method that can be used to reduce hallucinations in LLMs, as well as increase trust in AI-based tools by explicitly linking their output to external knowledge.

RAG involves augmenting LLMs with information retrieval (IR) systems, which can provide relevant content retrieved from external corpora as references. By incorporating external knowledge, retrieval-augmented LLMs can answer in-domain questions that cannot be answered by solely relying on the world knowledge stored in the model's own parameters.

Information retrieval systems often make use of pretrained models for Dense Passage Retrieval (DPR) and baseline systems such as BM25, a traditional IR strategy that uses term frequency-inverse document frequency (TF-IDF) weighting to rank documents based on their relevance to a query. In this work, we focus on creating a dense concept and sentence representation model which is well suited for use, among other tasks, as a DPR model within a biomedical RAG pipeline.

## Biomedical Representation Learning

To build a system capable of retrieving content relating to a concept by one of its names, the underlying models must be familiar with the vast biomedical terminology and the meaning of the underlying concepts. Because of the daunting scale of clinical terminology, in-domain pre-training is insufficient to cover long multi-word expressions accurately. Therefore, a now large body of work attempts to produce better representations using BKGs as a source, due to their extensive coverage of biomedical concepts.

Although multiple variations of the strategy exist, most state-of-the-art models prior to 2022 focused on learning representations of biomedical entities based on the synonyms of entities, in a contrastive manner. Since then, BioLORD-2022 was introduced to avoid spurious token-based overfitting using more of the relationships found in the knowledge graph, and by using full definitions to extract more fine-grained information out of medical knowledge bases. However, BioLORD-2022 did not make use of large language models to expand the training data, and it only used concept definitions during the pre-training phase. BioLORD-2022 models also suffer from a considerable performance decrease in general language understanding compared to the original pretrained language models they were based on, which is undesirable. In this work, we extend the BioLORD-2022 training strategy to fix these deficiencies.

## Cross-lingual Distillation of Knowledge

In addition to this, we provide a multilingual variant of our new model. While some multi-lingual biomedical models already exist, such as Multilingual SapBERT, their performance has been severely lagging behind the monolingual English models. This is because they rely on the existence of sufficient non-English training data during pre-training, but the existence of such data for the biomedical domain is scarce, and results in subpar generalization.

One of the goals of this work is to improve upon this state-of-the-art using proven techniques for cross lingual distillation [19]. Cross-lingual distillation trains a multilingual model to produce the same output as a target English model irrespective of the language in which a concept name is provided, such that "*Fever*", "*Fiebre*" and "*Fieber*" produce the same output.

# MATERIALS AND METHODS

In this work, we train a concept and sentence representation model finetuned for biomedical content, following a novel three-step strategy (see *Figure 1*) and evaluate its potential on several downstream tasks, all described in the following subsections. We also provide justifications for the changes made to the previous iterations of this strategy, with the use of ablation studies.

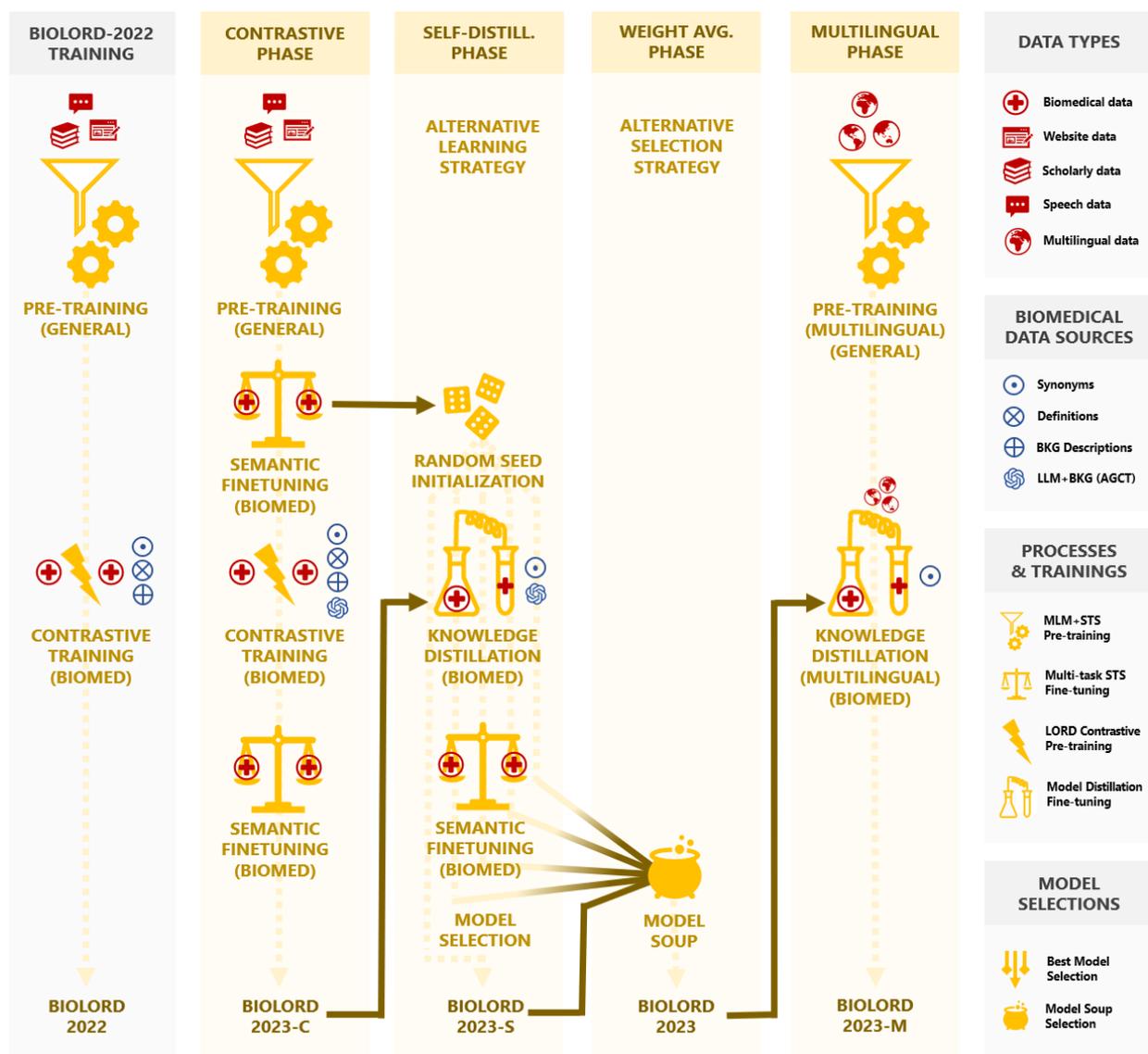

*Figure 1:* Compared to BioLORD-2022 (left), BioLORD-2023 involves a more advanced training strategy, composed of three phrases: a contrastive phase (further illustrated in Figure 2), a self-distillation phase (illustrated in Figure 3), and a weight-averaging phase (all further described in the subsections below).

## Training data

In addition to the algorithmic changes described in the next section, a core aspect of our work hinges on the new data we leverage during the training, and which we describe further in this section.

In addition to the training data already used in previous works, this study makes use of a large language model to generate a large set of definitions of biomedical concepts (grounded using the information contained in the SnomedCT ontology as context, as well as the information stored in the weights of the large language model itself). We leave the precise description of the procedure and its expert evaluation to the paper introducing the dataset [18], but we provide the most relevant details in the paragraph below.

In this new study, the pre-existing UMLS definitions are indeed complemented by 400,000 biomedical definitions from the Automatic Glossary of Clinical Terminology (AGCT), a large-scale biomedical dictionary of clinical concepts which we generated using the SnomedCT ontology and the GPT-3.5 language model. A subset of the generated definitions was evaluated by NLP researchers with biomedical expertise on three metrics: factuality, insight, and fluency; based on these metrics and a strict 6-grade quality rating, it was determined that more than 80% of the generated definitions would be usable for patient education, while more than 96% appeared useful for machine learning tasks. In this work, we set out to confirm whether that is truly the case in a practical scenario.

BioLORD-2023 also makes use of a newer version of the UMLS ontology (v2023AA) to generate its textual description, compared to BioLORD-2022 (which used v2020AB). This enables the new version of the model to become more aware of recent developments in the field, for example including knowledge related to the Covid19 pandemic.

# Training strategy

## Contrastive phase

To obtain our BioLORD-2023 model, we first make use of the contrastive objective devised by van den Oord et al. [36], with the goal of instilling biomedical knowledge into a base language model. More precisely, we make use of the "Learning Ontological Representations from Definitions" strategy (LORD) [11] in which batches of concept names and their definitions are fed to the language model, and where the distance between the representation of a concept name and the representation of its definition should be minimized, while maximizing the distance between a concept and the definition of the other concepts in the batch.

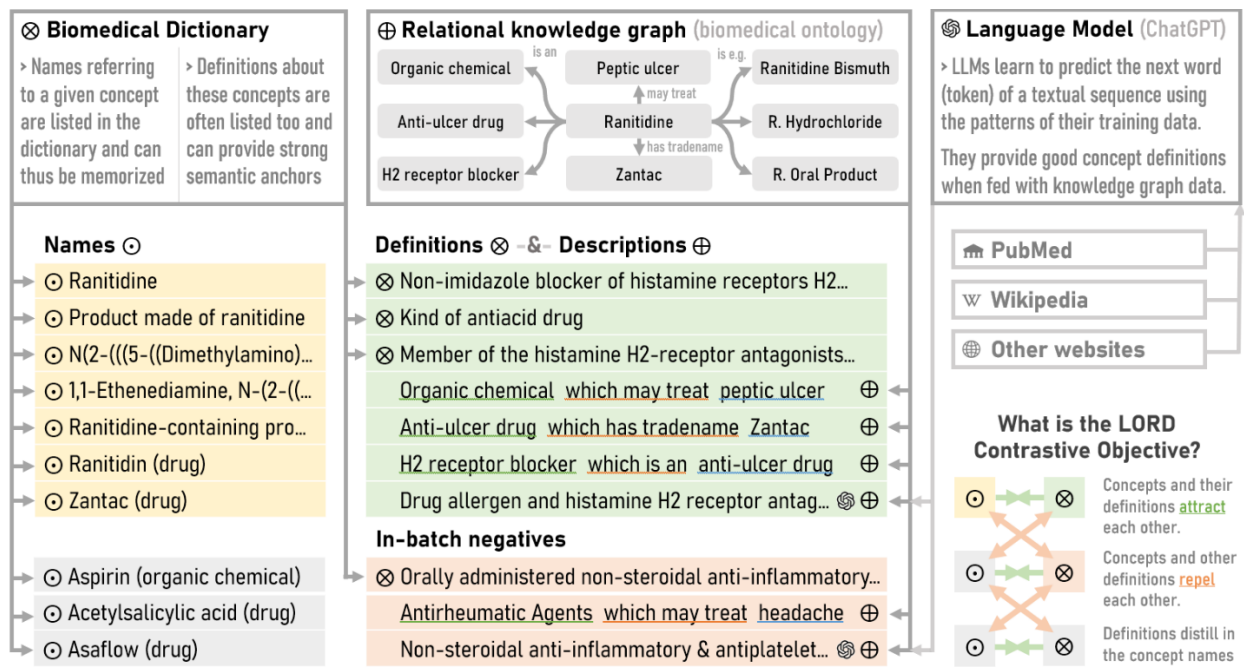

*Figure 2:* BioLORD aims to bring the representation of biomedical concept names (⊙) and their definitions (⊗) closer to each other, to ground the name representations with knowledge from the definitions. This is illustrated for the Ranitidine and Aspirin concepts from UMLS. Knowledge from the ontology's relational knowledge graph is injected by extending the set of known definitions with automatically generated definitions (⊕) from the Automatic🅢 Glossary of Clinical Terminology, as well as with simpler template-based descriptions sampled from UMLS relationships (⊕). Contrastive learning is applied to attract the representations of compatible pairs (⊙, ⊗ or ⊕) and repel incompatible ones.

For instance, the representations by the language model of "ranitidine" and of "an H2-antagonist substance frequently used to treat peptic ulcer" [16] are encouraged to be as close as possible, while remaining far from the representations of "aspirin" and of "a synthetic compound used medicinally to relieve mild or chronic pain and to reduce fever and inflammation" [16], which are related to another concept of the batch.

In this work, we make use of the improved initialization strategy developed by Remy et al [37], which involves adapting the STAMB2 model [38] used for the initialization of BioLORD-2022 to the STS tasks of our benchmark in a multi-task setup, prior to applying our contrastive phase. Multi-task setups have been shown to be effective in scenarios where catastrophic forgetting is possible as a result of continual learning [39]. By aligning the STAMB2 model to the human preference of STS datasets prior to applying the contrastive learning phase of BioLORD, BioLORD-2023 produces representations of medical concepts which align more closely with human judgment.

We also insert this adaptation phase a second time, after the contrastive learning stage of BioLORD. This further enhances the BioLORD-2023 model's performance on STS tasks.

**Self-distillation phase**

The contrastive pre-training strategy described above was shown to yield excellent results in biomedical knowledge acquisition. We briefly summarize below the findings of BioLORD-2022 [11] and refer to the paper for details.

Compared to their STAMB2 base [38], models trained using the BioLORD-2022 methodology show a visible improvement in clinical sentence understanding (from 85.9 to 86.3 Pearson correlation scores on MedSTS) and greatly improved biomedical concept representation capabilities (from 46.7 to 56.0 on UMNSRS).

Unlike models trained using the SapBERT training strategy, BioLORD-2022 models remained proficient in the handling of sentence-level semantics (89.3% on SICK, compared to only 80.3% for SapBERT). We attribute this to our choice of model initialization and the inclusion of concept definitions in the BioLORD-2022 training strategy, which succeeded in preventing a catastrophic forgetting of sentence parsing during the biomedical knowledge acquisition phase.

However, the addition of the definitions did not prove sufficient to avoid a measurable degradation of the performance of the model in general-purpose semantic similarity tasks (from 88.0 down to 86.5 Pearson correlation score on STS-Benchmark, comparing STAMB2 to BioLORD-2022). This degradation cannot be solely attributed to a loss of knowledge about general-domain concepts, as the performance degradation remained visible even when no general-domain knowledge was required for solving the semantic task (from 90.7 down to 89.3 Pearson correlation score on SICK, comparing STAMB2 to BioLORD-2022).

We attribute a large part of this performance degradation to the semantic space distortion induced by the extensive contrastive learning on concept names and their definitions, which only elicits a select few aspects of the semantic textual similarity task, blowing up the importance of these aspects considerably at the expense of other aspects of language understanding, thereby resulting in a loss of calibration of the model.

To counter-balance this, we propose to substitute the unsupervised contrastive learning phase by a supervised objective, taking into account the learnings of the contrastive phase without having to resort to a contrastive objective. To this end, we generate concept embeddings for the 4 million biomedical concepts contained in UMLS (using the BioLORD-2023-C model), and finetune a base-model (which has not undergone the contrastive learning phase) to accurately predict these distilled concept embeddings via a learned projection followed by a standard Mean-Squared-Error loss (see *Figure 3*).

We call this process the "self-distillation phase", as we distill the knowledge acquired by the contrastive model into a past version of itself in a supervised manner, hoping that this will better preserve its existing knowledge.

Like in the previous phase, we leverage the knowledge extracted from the knowledge graphs and the LLM by incorporating the concept definitions in the distillation process. We do this by first producing embeddings for concept names and their generated concept definitions using the model resulting from the contrastive phase of BioLORD-2023, and by subsequently averaging these two representations for each concept and using the result as the regression target during self-distillation, for both the concept name and its definition.

To improve the training speed, we reduce the latent space of produced embeddings to 64 dimensions through PCA. Finally, we train a randomly-initialized linear projection head on top of the base STAMB2-STS model to predict these 64-dimension embeddings (see *Figure 3* below).

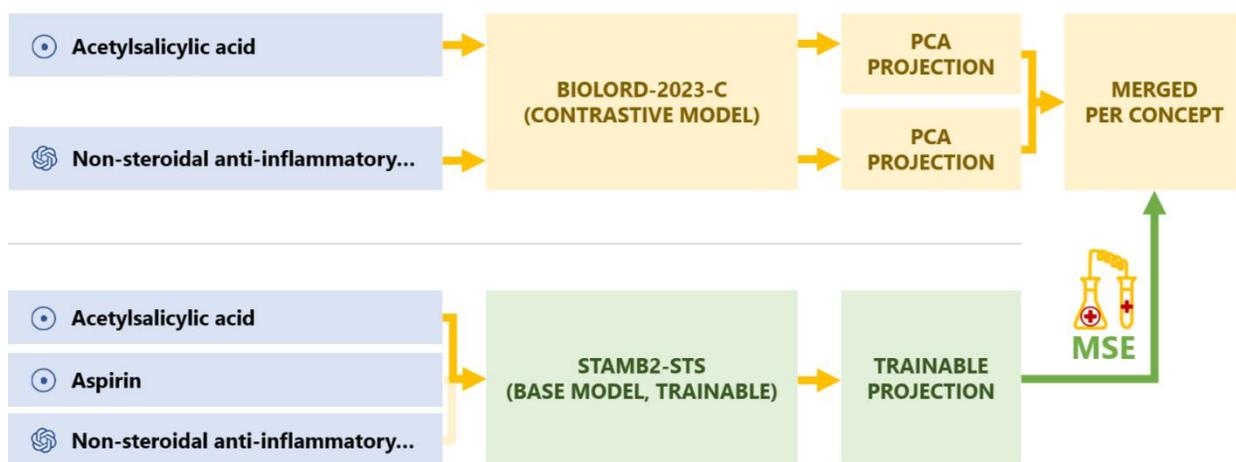

*Figure 3:* In the self-distillation phase, the knowledge acquired during the contrastive phase is imbued into the base model, using a more direct training strategy. The representation of each textual variant of a concept is trained to map the average of the contrastive model representation of its name and definition.

This supervised self-distillation phase possesses several key advantages: <1> by including the concept definitions in the process to produce the concept embeddings, better representations can be learnt for biomedical concepts whose name is otherwise uninformative or difficult for the model to memorize, (2) it is considerably faster than the contrastive learning phase, which is likely to cause less forgetting of the original task for an identical level of new knowledge acquisition [40] (3) it further enables the language model to leverage its existing features to obtain the desired knowledge, without having to distort them to reduce the in-domain anisotropy [41-43], and (4) it should be possible to apply this distillation phase on a different base model than the one used during the contrastive learning phase, taking advantage of improved base models at a limited training cost (we however leave the exploration of this aspect for future works).

**Weight-averaging phase**

An interesting aspect of the self-distillation phase described above is that it hinges on a randomly-initialized projection head added on top of the base sentence representation model. As a result of this, different random seeds result in slightly different models, focusing on different aspects of the sentence embedding.

While a commonly used technique in this scenario, called hyperparameter tuning, aims to select the best model from multiple experiments based on a held-out validation set, Wortsman et al. [44] discovered a better strategy, which they named "model soups", and which consists in the averaging of the weight of the parameters of several finetuned models.

Indeed, Wortsman and his colleagues were able to show that it is often possible to improve the accuracy and robustness of the resulting system by averaging the weights of multiple models, each fine-tuned with different hyperparameter configurations. Unlike a traditional ensemble, no additional inference or memory costs are incurred as a result of the merge, irrespective of the number of models being merged, which makes this technique particularly attractive for DPR models, where a fast inference is highly desirable. We evaluate the impact of the weight averaging in our Discussion Section.

**Cross-lingual distillation**

Finally, we further make use in this work of the technique described by Reimers et al. [19], which consists in using a parallel corpus to distill a high-quality monolingual model into a multi-lingual model yielding similar representations for a text in that language and translations sourced from the corpus.

We cross-lingually distill the representations of our English model into the "paraphrase-multilingual-mpnet-base-v2" language model introduced in the same paper as the distillation technique, and which supports 50+ languages.

In particular, we make use of an aligned corpus generated from the regional releases of SnomedCT, using the Google LaBSE bi-text mining model [20], and which we describe in more detail in our EmP 2022 publication [45]. This aligned corpus contains alignments for the following languages: English, Spanish, French, German, Dutch, Danish, and Swedish.

While we did not investigate this in the current version of BioLORD-2023-M, this corpus could be complemented by multilingual annotations from UMLS (and in particular its MESH subset), to increase language coverage. We leave this investigation to a follow-up work.

# Evaluation methodology

As mentioned in the abstract, an extensive test suite is required to evaluate the capabilities of semantic models. The following paragraphs list and describe the various benchmarks used to evaluate BioLORD-2023, covering semantic textual similarity, concept representations, and entity linking, all in the biomedical and clinical domains.

## Clinical Semantic Textual Similarity

Semantic Textual Similarity is an NLP task measuring the degree of semantic alignment between NLP models and human judgment, by assigning similarity scores to a pair of two sentences, usually from 0 to 5, and computing the correlation between scores obtained by expert human judgement and model-assigned scores.

We evaluate the STS capabilities of the models on five popular benchmarks: three biomedical or clinical ones (MedSTS [46], MedNLI-S [47], BIOSSES [48]), and two general purposes benchmarks (SICK [49] and STS-Benchmark [50]). For readers unfamiliar with these datasets, we provide a brief introduction for each of them in Appendix C.

## Biomedical Concept Representation

Known as Biomedical Concept Representation, this task concerns the mapping of biomedical concepts to a vector latent space, whose features enable classifying these concepts, or deriving properties from them. It can be relevant for numerous biomedical tasks including disease subtype annotations [51].

Following the approach of Kalyan and Sangeetha [52], we evaluate our model using four benchmarks: EHR-RelB [53], UMNSRS-Similarity [54], UMNSRS-Relatedness [54], and MayoSRS [55]. For readers unfamiliar with these datasets, we provide a brief introduction for each of them in Appendix C.

**Biomedical Named Entity Linking**

The task of biomedical concept name normalization (BCNN), also referred to as Named Entity Linking (NEL) in the broader literature, concerns the mapping of free-from text describing clinical disorders or concepts to a fixed list of biomedical concepts, such as the elements of biomedical ontology.

To showcase improvements in NEL, we reuse the evaluation setup devised by Portelli et al. [56], where biomedical language models were evaluated on a set of five datasets of varying levels of formality, listed here in the reverse order of formality (least formal first): TwiMed-Twitter [57], SMM4H [58], PsyTar [59], CADEC [60], and TwiMed-PubMed [57]. For readers unfamiliar with these datasets, we provide a brief introduction for each of them in Appendix C.

## Hyperparameters and training details

In order to focus on the effect of the inclusion of LLM-generated definitions in the training set, and of our improve training strategy including the novel self-distillation step, all the experiments that follow are finetuned from the same base model as BioLORD-2022.

This base model had a size identical to the other baselines models evaluated in this study, enabling a fair comparison between them. We also report some interesting findings about larger models in Appendix A.

To facilitate the replication of our results, we release the code used in the various phases of our training jointly with this paper, and detail our choices of hyperparameters in Appendix B.

# RESULTS

This section presents the empirical evaluation results of BioLORD-2023, in comparison to existing models. In order to gain insights into the modified training strategy, compared to BioLORD-2022, a number of ablation results are provided as well. Finally, our multi-lingual model is also evaluated.

For our evaluation of the English BioLORD-2023, we follow a structure similar to BioLORD-2022, and analyze in turn the suitability of the model for several tasks including Clinical Semantic Textual Similarity, Biomedical Concept Representation, and Named Entity Linking (as described in the previous section).

We report these results in Table 1.

|     |              | BioSyn [9] | SapBERT [10] | BioLORD-2022 | BioLORD-2023 |
|-----|--------------|------------|--------------|--------------|--------------|
| STS | MedSTS [46]  | 84.0       | 86.0         | 86.3 🥈      | 88.3 🥇      |
|     | MedNLI-S [47]| 89.5       | 90.5 🥈      | 89.9         | 92.4 🥇      |
|     | BIOSSES [48] | 92.1 🥇    | 89.3 🥈      | 84.0         | 86.1         |
|     | SICK [49]    | 86.7       | 80.3         | 89.3 🥈      | 90.3 🥇      |
|     | STS [50]     | 79.4       | 81.9         | 86.5 🥈      | 87.8 🥇      |
| BCR | EHR-Rel-B [53]| 42.5      | 51.7         | 57.5 🥈      | 63.6 🥇      |
|     | UMNSRS-S [54]| 43.6       | 53.0         | 56.0 🥈      | 59.2 🥇      |
|     | UMNSRS-R [54]| 39.1       | 47.5         | 54.4 🥇      | 54.4 🥇      |
|     | MayoSRS-S [55]| 45.1      | 62.5         | 74.7 🥇      | 74.4 🥈      |
| NEL | TwiMed-TW [57]| 42.8      | 48.3         | 48.5 🥈      | 49.8 🥇      |
|     | SMM4H [58]   | 33.1       | 43.4         | 46.5 🥈      | 47.7 🥇      |
|     | PsyTAR [59]  | 52.4       | 64.8 🥈      | 64.7         | 66.3 🥇      |
|     | CADEC [60]   | 35.3       | 40.4         | 58.7 🥈      | 63.0 🥇      |
|     | TwiMed-PM [57]| 65.3      | 70.1 🥈      | 70.4 🥇      | 69.4         |

**Table 1:** Performance characteristics of state-of-the-art biomedical models on **STS** (Pearson correlation), **BCR** (Spearman correlation), and **NEL** (Top1 Accuracy). The following models are evaluated: **BioSyn** (state-of-the-art in 2020), **SapBERT** (state-of-the-art in 2021), **BioLORD-2022** (our baseline), and **BioLORD-2023** (our new model). We underline the best and second-best models with a gold and/or silver medal, respectively.

We also conduct an ablation study, showing the effect of the various training phases of the BioLORD-2023 methodology. To compare the effect of the training strategies more effectively, we report the absolute improvements over the base model (STAMB2 in all cases, for BioLORD-2022 and 2023 models).

Results are shown in Table 2.

|  |  | STAMB2 [38] (base) | BioLORD-2022 | BioLORD-2023-C | BioLORD-2023-S | BioLORD-2023 |
|---|---|---|---|---|---|---|
| STS | MedSTS [46] | 85.9 | +0.4 | +0.4 | +1.6 🥈 | +2.4 🥇 |
|  | MedNLI-S [47] | 89.4 | +0.5 | +2.5 | +2.7 🥈 | +3.0 🥇 |
|  | BIOSSES [48] | 90.7 | −6.7 | −4.5 🥇 | −5.3 | −4.6 🥈 |
|  | SICK [49] | 90.7 | −1.4 | −1.0 | −0.8 🥈 | −0.4 🥇 |
|  | STS [50] | 88.0 | −1.5 | −1.1 | −1.0 🥈 | −0.2 🥇 |
| BCR | EHR-Rel-B [53] | 47.1 | +10.4 | +11.2 | +15.7 | +16.5 🥇 |
|  | UMNSRS-S [54] | 43.9 | +12.1 | +13.7 | +15.3 🥇 | +15.3 🥇 |
|  | UMNSRS-R [54] | 46.7 | + 7.7 | + 8.0 🥈 | + 9.1 🥇 | + 7.7 |
|  | MayoSRS-S [55] | 54.1 | +20.6 🥇 | +18.6 | +15.9 | +20.3 🥈 |
| NEL | TwiMed-TW [57] | 44.1 | + 4.4 | + 5.7 🥇 | + 3.7 | + 5.7 🥇 |
|  | SMM4H [58] | 38.4 | + 8.1 | + 8.5 🥈 | + 3.2 | + 9.3 🥇 |
|  | PsyTAR [59] | 56.5 | + 8.2 | + 8.9 🥈 | + 5.2 | + 9.8 🥇 |
|  | CADEC [60] | 36.9 | +21.8 | +26.2 🥇 | +24.4 | +26.1 🥈 |
|  | TwiMed-PM [57] | 62.8 | + 7.6 🥇 | + 4.9 | + 5.7 | + 6.6 🥈 |

**Table 2:** Performance characteristics of the BioLORD models obtained after each proposed training phase, relative to the STAMB2 performance (in percentage points), on **STS** (Pearson correlation), **BCR** (Spearman correlation), and **NEL** (Top1 Accuracy). The following models are evaluated: **STAMB2** (our shared base model), **BioLORD-2022** (our baseline), **BioLORD-2023** (our new model), and an ablation study for each intermediary training phase (**BioLORD-2023-C** for the contrastive phase, and **BioLORD-2023-S** for the Self-Distillation phase; see Figure 1). We underline the best and second-best models with a gold and/or silver medal, respectively.

In addition, we separately evaluate our cross-lingual model using both the English test suite (to evaluate the distillation quality) and a multilingual named entity linking task, XL-BEL.

For XL-BEL, we report the 3 European languages on which both multilingual SapBERT and BioLORD-2023-M were finetuned: German, Spanish and English.

We report those results in Table 3.

|  |  | SapBERT [10] | mSapBERT [21] | BioLORD-2023-M | BioLORD-2023 |
|---|---|---|---|---|---|
| STS | MedSTS [46] | 86.0 | 85.6 | 86.0 🥇 | 88.3 |
| | MedNLI-S [47] | 90.5 | 88.1 | 92.1 🥇 | 92.4 |
| | BIOSSES [48] | 89.3 | 90.0 🥇 | 75.4 | 86.1 |
| | SICK [49] | 80.3 | 87.0 | 89.1 🥇 | 90.3 |
| | STS [50] | 81.9 | 83.5 | 85.1 🥇 | 87.8 |
| BCR | EHR-Rel-B [53] | 51.7 | 42.4 | 64.1 🥇 | 63.6 |
| | UMNSRS-S [54] | 53.0 | 34.2 | 60.1 🥇 | 59.2 |
| | UMNSRS-R [54] | 47.5 | 29.6 | 54.3 🥇 | 54.4 |
| | MayoSRS-S [55] | 62.5 | 45.2 | 74.8 🥇 | 74.4 |
| NEL | TwiMed-TW [57] | 48.3 | 47.4 | 49.3 🥇 | 47.4 |
| | SMM4H [58] | 43.4 | 40.8 | 42.3 🥇 | 46.9 |
| | PsyTAR [59] | 64.8 | 51.5 | 63.3 🥇 | 66.3 |
| | CADEC [60] | 40.4 | 46.8 | 47.0 🥇 | 47.4 |
| | TwiMed-PM [57] | 70.1 | 63.9 | 67.4 🥇 | 69.4 |
| MNEL | German XLB [21] | N/A | 51.5 | 57.7 🥇 | N/A |
| | Spanish XLB [21] | N/A | 52.7 | 53.1 🥇 | N/A |
| | English XLB [21] | N/A | 78.2 🥇 | 73.1 | N/A |

*Table 3:* Performance characteristics of state-of-the-art multilingual biomedical models on **STS** (Pearson correlation), **BCR** (Spearman correlation), **NEL** (Top1 Accuracy), and **Multilingual NEL** (Top1 Accuracy). We underline the best multilingual model with a gold medal (English-only models are only provided for comparison purposes).

# DISCUSSION

This section provides insights into the results, structured according to the benchmark tasks, covering the absolute metrics reported in Table 1 and the impact of training strategies as reported in Table 2. After discussing the results for the English BioLORD-2023, we will demonstrate the quality of its multi-lingual counterpart, BioLORD-2023-M, by referring to Table 3.

## Clinical Semantic Textual Similarity

As can be seen in Table 1, our new model BioLORD-2023 demonstrates a considerably increased performance on the biomedical tasks such as MedSTS (from 86.3 to 88.3), BIOSSES (from 84.0 to 86.1) and MedNLI-S (from 89.9 to 92.4) while also increasing its performance in general-purpose tasks like the STS-Benchmark (from 86.5 to 87.8) and SICK (from 89.3 to 90.3).

These gains in the general-purpose domain stem from a vastly reduced performance degradation from the original STAMB2 model (from 88.0 down to 87.8 for STS-B, and from 90.7 to 90.3 for SICK). Given that our BioLORD training strategy does not focus on the general domain, we cannot expect any performance improvement on these tasks. In fact, such a minor degradation can be considered a success. Indeed, incurring some performance degradation is inevitable as the finetuned model accumulates more biomedical knowledge (this remains the main objective of the BioLORD approach).

The performance of biomedical semantic models on the general domain was shown in previous studies [37] to be a good indicator of the NEL performance of the models in less formal contexts, such as healthcare information posted on social media, a crucial information source for pharmacovigilance. We also believe that it is a good indication of model robustness, as not all clinical notes are written in formal and unambiguous medical language.

The above results demonstrate with confidence that BioLORD-2023 is the new state-of-the-art semantic model for the biomedical domain.

## Biomedical Concept Representation

In this section, we analyze the performance of the embeddings produced by state-of-the-art models for the task of modelling biomedical concepts.

BioLORD-2023 obtains superior performance on all the considered BCR benchmarks, as reported in Table 1. It performs particularly well on the EHR-Rel-B benchmark (the most exhaustive and recent one) as well as the UMNSRS-Similarity benchmark.

The ablation results of Table 2 show that the evolution of the scores over the training phase follows a similar pattern to the clinical STS tasks already presented in the previous subsection, although there appears to be some trade-off between Similarity tasks and Relatedness tasks.

This time again, the results are consistent with our hypothesis that BioLORD-2023 is the current best biomedical embedder, even beating more complex systems that explicitly combine graphs and text embedders such as Kalyan and Sangeetha [52] and Mao and Fung [61]. We refer to these papers for more detail on their results and methodology.

## Biomedical Named Entity Linking

Overall, the strong results for the biomedical NEL tasks confirm that BioLORD-2023 performs well both on more formal datasets such as TwiMed-PM and on informal medical datasets, such as TwiMed-TW.

Unlike the STS and BCR tasks, we do not notice a similarly clear pattern of decreased performance for the ablation studies. We suspect that NEL is a task that favors heavily contrastive learning strategies, making the gains of the self-distillation less relevant.

## Cross-lingual distillation

In this section, we evaluate the performance of the multilingual BioLORD model (referred to as BioLORD-2023-M), and further discuss its potential.

We evaluate BioLORD-2023-M in two ways. Firstly, we evaluate its performance on the same datasets used for our English evaluation, to determine the impact of the multilingual distillation on the model performance. Secondly, we assess its concept name normalization capabilities for languages other than English, such as Spanish and German, using the procedure used by Multilingual SapBERT.

In Table 3, we report the performance of multilingual variants of SapBERT and BioLORD on the English tasks on which their monolingual equivalents were evaluated before.

As a result of the distillation procedure, which focuses on biomedical concept names, BioLORD-2023-M achieves comparable or superior performance on BCR benchmarks compared to BioLORD-2023. When it comes to sentence-level STS tasks, some performance degradation is incurred, but BioLORD-2023-M remains well ahead of its SapBERT competitors. These results indicate that the multilingual distillation procedure was highly effective and displays strong performance on English tasks for which the original model excelled.

When we consider the performance of these multilingual models on NEL tasks in English. This time again, we find comparable performances between the original and the distilled model, albeit the monolingual model usually performs slightly better. In all cases, BioLORD-2023-M performs considerably better than multilingual SapBERT on these tasks.

Finally, we also report in Table 3 the performance of multilingual SapBERT and multilingual BioLORD on the clinical subset of XL-BEL, a dataset specifically developed for the evaluation of multilingual SapBERT. Because we noted that a large proportion of the evaluation set concerns the names of plant and animal species, which are not very relevant to the clinical setting these models were developed for, we filtered the XL-BEL test set to exclude these mention types (based on their UMLS Semantic Types). This enables the evaluation to focus on all the other semantic types, such as clinical disorders, drugs, and procedures.

Overall, BioLORD-2023-M achieves better results compared to multilingual SapBERT on both German and Spanish, the two non-English languages supported by both models. Interestingly, multilingual SapBERT keeps an edge when it comes to English data, but monolingual models would be better suited for this task than a multilingual model.

In summary, the results of this section demonstrate that our multilingual BioLORD-2023-M model achieves comparable performance on multilingual NEL as the multilingual SapBERT model, while achieving considerably better results on the English STS and BCR tasks than both the English and Multilingual SapBERT on these datasets. This makes our multilingual model a solid choice for a large range of biomedical tasks in the supported languages.

## Expected impact

With its better semantic understanding of the biomedical domain, BioLORD-2023 holds the potential to greatly enhance our comprehension of diseases, helping researchers to decipher complex biomedical literature, uncovering relationships between genes, proteins, and pathways, and thereby accelerating the pace of biomedical research and drug discovery.

Moreover, our model enables a nuanced understanding of patient records, extracting meaningful insights that may have otherwise remained buried. This better comprehension of clinical narratives can contribute to more accurate diagnoses, personalized treatment plans, and improved patient outcomes.

# CONCLUSION

In this study, we introduced BioLORD-2023, a model offering state-of-the-art capabilities for clinical semantic textual similarity and biomedical concept representation.

Through the introduction of innovative techniques such as the inclusion of LLM-generated definitions in the training data, a supervised self-distillation phase, a robust model-weights averaging, and a state-of-the-art cross-lingual distillation, we were able to train a series of models which confidently demonstrates substantial performance improvements over their predecessors across a wide range of tasks.

These enhancements have real-world implications, making BioLORD-2023 a valuable tool for healthcare professionals, researchers, and patients, thereby contributing to the ongoing progress in the field of biomedical natural language processing.

To support the research community, we release our trained models for all three stages of our pipeline on HuggingFace. This will also enable other researchers to pick the model that is best suited for their experiments.

https://huggingface.co/FremyCompany/BioLORD-2023

# LIMITATIONS AND FUTURE WORK

While our work showcases an impressive improvement over the previous state of the art, it is not without limitations. One such limitation is the bound of its knowledge, which is constrained by the knowledge contained in the knowledge graphs used to prompt the language model.

Many entities in the knowledge graph contain few pieces of information, and this is particularly the case for organisms and species, making BioLORD models inadequate to differentiate between such entities, as they acquire very similar representations. Another limitation is that knowledge graphs trail the literature and might not include all new pieces of knowledge mentioned in the published biomedical papers.

To address this problem, we foresee a future version of this model which would rely on a combination of the knowledge graph information and of the recent biomedical literature to prompt LLMs before generating definitions, thereby including even more recent and relevant information, and helping create a more precise definitions for rare or very specific concepts.

The retrieval of the relevant documents to include in the prompt will already benefit from our state-of-the-art BioLORD-2023 model. We hope to study the impact of grounding document retrieval on the generated definitions in an upcoming study.

We also foresee future versions of BioLORD or similar semantic models making use of the larger STS models that are starting to become available. While most of the successful models remain closed-sourced so far and can therefore not be freely finetuned for the biomedical domain, new and larger models are expected to be open-sourced as time goes by. Applying the self-distillation phase of BioLORD to these models could prove an efficient way to leverage them.

# ACKNOWLEDGEMENTS

This work would not have been possible without the joint financial support of the Vlaams Agentschap Innoveren & Ondernemen (VLAIO) and the RADar innovation center of the AZ Delta hospital group, through their ambitious joint-venture in the Advanced Data-Aided Medicine project (ADAM).

I would of course like to extend special thanks to Karel D'Oosterlinck, Alfiya Khabibullina, Beatrice Portelli, Simone Scarboro, Peter De Jaeger, and other individuals who helped me make progress along the way towards this journal publication, often in the form of joint conference or workshop publications.

A particular thanks is also extended to the team maintaining and developing GPULab, the machine learning infrastructure for AI computing built in collaboration between UGent, UAntwerpen and the imec research and development center.

Finally, I would also like to thank my co-supervisors, Thomas Demeester and Kris Demuynck, for their support and constructive advice during the ideation process, and all along the development of this project up to this very article.

# APPENDIX A: NEGATIVE RESULTS

In this section, we report on alternative designs that we considered while developing our BioLORD pipeline. These results provide some insights into the limitations of our approach and suggest directions for future work.

- We initially hypothesized that generating definitions for the concepts using the knowledge graph and LLMs would replace the need for textual descriptions sampled from the knowledge graph. However, our experiments showed that the textual descriptions still contributed positively to the performance of our pipeline, as they enforced an attraction between the concepts and their parent concept names (which we used in the template). Therefore, we decided to use both the definitions and the textual descriptions in our final pipeline.

- We also investigated the effect of using different biomedical models as base models for our BioLORD models. We expected that these models would have an advantage over general-purpose language models fine-tuned on STS, as they were pre-trained on large-scale biomedical corpora. However, our results did not confirm this expectation, as biomedical models performed worse than the fine-tuned models. We attribute this to a lack of sentence understanding in the biomedical models, which led them to misunderstand definitions and overfit on shared tokens instead of semantic similarity. Thus, we concluded that base models with STS pre-training are critical for producing BioLORD models.

- Finally, we attempted to include hard mining in our pipeline. We implemented hard mining based on the ontological knowledge graph, by choosing siblings or siblings of ancestors (after ensuring they were not ancestors of the current concept themselves) as hard negatives. While hard mining improved NEL, it degraded semantics; it seems that hard mining was too aggressive and penalized concepts that were not very dissimilar. We think that a more sophisticated approach than hard mining would be needed in this case, with a margin which depends on the two concepts being compared.

# APPENDIX B: DETAILS ON OUR EXPERIMENTAL SETUPS

In this section, we summarize the details of the experiments described in the results section. These details are mainly aimed at helping other researchers willing to replicate our experiments, but they might be insightful on their own to understand the scale of the work.

## Contrastive phase

We leveraged our recently-upgraded hardware and ran our new contrastive phase experiments on an NVIDIA A40 GPU with 48Gb memory, up from the 32 Gb of the V100 GPU used in the BioLORD-2022 experiments. As a result, we increased the batch size from 96 to 128. We let our experiments run for about 9 days, corresponding to one epoch over our combined dataset.

This duration is very close to our previous experiments (7 days) and is the result of a careful balance between the above-mentioned hardware upgrade and our increased demand for flops. We reuse the best-performing hyperparameters from our past experiments (AdamW for the optimizer, WarmupLinear for the scheduler, 2e-5 for the learning rate, 5% of the data for the warmup window, 0.01 for the weight decay, 1 for the number of epochs, PyTorch 2.0 AMP for the mixed-precision training).

We release the code of our unsupervised contrastive training for easier replication of our results.

## Self-distillation phase

We ran our self-distillation experiments on a single NVIDIA 3090 GPU, given the low memory requirements and relatively short training time required for this operation. Each training run takes about 5 hours (1 hour per epoch), with the default hyperparameters settings (AdamW for the optimizer, WarmupLinear for the scheduler, 2e-5 for the learning rate, 5% of the data for the warmup window, 0.01 for the weight decay, 5 for the number of epochs, PyTorch 1.7.1 AMP for the mixed-precision training). We also release the code of our supervised training to facilitate the replication of our results.

## Weight averaging phase

The weight averaging phase implements the Greedy Model Soup strategy described in the original paper [44]. In our experiment, we found that merging 7 closely-related fine-tuned models performed best among the configurations which were attempted. We release the resulting average weights as our final BioLORD-2023 model.

## Cross-Lingual distillation

We ran our cross-lingual distillation experiments on a single NVIDIA 3090 GPU, given the low memory requirements and relatively short training time required for this operation. Each training run takes about 40 hours (4h per epoch), with the default hyperparameters settings (AdamW for the optimizer, WarmupLinear for the scheduler, 2e-5 for the learning rate, 5% of the data for the warmup window, 0.01 for the weight decay, 10 for the number of epochs, PyTorch 1.7.1 AMP for the mixed-precision training).

The training data consists of all UMLS concept names, as well as parallel translation pairs for SnomedCT terms in the regional languages it supports (i.e. 1428k pairs in Spanish, 703k in French, 452k in German, 1444k in Dutch, 284k in Danish, and 412k in Swedish).

We leave for a future work the usage of UMLS synonyms in languages other than English, and the addition of definitions in the distillation procedure.

# APPENDIX C: DESCRIPTION OF EVALUATION DATASETS

## Clinical Semantic Textual Similarity

**MedSTS** [46] is a dataset which was developed for evaluating clinical semantic textual similarity. It contains 1,068 sentence pairs which were annotated by two medical experts with semantic similarity scores of 0-5 (low to high similarity).

**MedNLI** [47] is a dataset initially developed for evaluating natural language inference reasoning over the clinical domain. It is curated by doctors tasked with providing three statements (one entailed, one contradicted, and one neutral) grounded in the medical history of a given patient. We report the proportion of hypothesis statements which are more similar to their entailed statement than their contradictory statement.

**BIOSSES** [48] is a biomedical semantic similarity dataset containing 100 sentence pairs and which focuses on scientific articles in the biomedical domain, rather than clinical notes. It is a challenging dataset because of the length of its entries, which often contain several sub-sentences.

**SICK** [49] is a dataset which consists of about 10k English sentence pairs, designed to be rich in lexical, syntactic, and semantic phenomena. Pairs have been annotated for relatedness on a 0-5 scale.

**STS Benchmark** [50] is a dataset which regroups several other general-purpose text similarity datasets (and contains 8628 sentence pairs). It was developed as a public benchmark for the first shared task of SemEval-2017, a workshop focusing on the evaluation of semantic models.

## Biomedical Concept Representation

**EHR-RelB** [53] is a dataset containing 3630 concept pairs sampled from electronic health records, rated for relatedness by 3 doctors.

**UMNSRS** [54] is a pair of datasets, consisting of 725 clinical term pairs whose semantic similarity and relatedness were determined on a continuous scale by 4 clinicians.

**MayoSRS** [55] is a dataset formed by 101 clinical term pairs whose relatedness was reported on a 4-point scale by nine medical coders and three physicians.

## Biomedical Named Entity Linking

**TwiMed** [57] provides a comparable corpus of texts from PubMed (abstracts) and Twitter (posts), allowing pharmacovigilance researchers to better understand the similarities and differences between the language used to describe disease and drug-related symptoms on PubMed (TwiMed-PM, clinical domain) and Twitter (TwiMed-TW, social media domain). Both sets of data contain 1000 samples.

**SMM4H** [58] is a dataset for Adverse Drug Event (ADE) normalization. It was used in the SMM4H 2020 shared task on ADE normalization. The aim of the subtask was to recognize ADE mentions from tweets and normalize them to their preferred term in the MedDRA ontology. The dataset includes 1212 tweets.

**PsyTAR** [59] contains patients' expression of effectiveness and adverse drug events associated with psychiatric medications, originating from a sample of 891 drugs reviews posted by patients on an online healthcare forum.

**CADEC** [60] is a corpus of user-generated reviews of drugs that has been annotated with adverse drug events (ADEs) and their normalization. It contains 1250 posts from a medical forum, which were annotated by a team of experts from the University of Arizona.